\documentclass[runningheads]{llncs}
\usepackage{graphicx}
\usepackage{tikz}
\usepackage{comment}
\usepackage{amsmath,amssymb} % define this before the line numbering.
\usepackage{color}
\usepackage{xcolor}
\newcommand{\tcl}{\textcolor{red}}
\newcommand{\egno}{\textit{e.g.}}
\newcommand{\ieno}{\textit{i.e.}}
\usepackage{multirow}
\usepackage{booktabs}
\usepackage{subfig}
\usepackage[colorlinks,urlcolor=magenta]{hyperref}
\usepackage[accsupp]{axessibility}  % Improves PDF readability for those with disabilities.

\begin{document}
% \linenumbers
\pagestyle{headings}
\mainmatter
\def\ECCVSubNumber{64}  % Insert your submission number here

\title{HST: Hierarchical Swin Transformer for Compressed Image Super-resolution} % Replace with your title

\begin{comment}
\titlerunning{ECCV-22 submission ID \ECCVSubNumber} 
\authorrunning{ECCV-22 submission ID \ECCVSubNumber} 
\author{Anonymous ECCV submission}
\institute{Paper ID \ECCVSubNumber}
\end{comment}
%******************

%\begin{comment}
\titlerunning{HST}
%\orcidID{1111-2222-3333-4444}
\author{Bingchen Li \and Xin Li \and
Yiting Lu \and Sen Liu \and Ruoyu Feng \and Zhibo Chen\thanks{corresponding author}}
\authorrunning{B. Li et al.}
\institute{University of Science and Technology of China, Hefei 230027, China\\
\email{\{lbc31415926, lixin666, luyt31415\}@mail.ustc.edu.cn, elsen@iat.ustc.edu.cn, ustcfry@mail.ustc.edu.cn, chenzhibo@ustc.edu.cn} \\}
%\end{comment}
%******************
\maketitle

\begin{abstract}
Compressed Image Super-resolution has achieved great attention in recent years, where images are degraded with compression artifacts and low-resolution artifacts. Since the complex hybrid distortions, it is hard to restore the distorted image with the simple cooperation of super-resolution and compression artifacts removing. In this paper, we take a step forward to propose the Hierarchical Swin Transformer (HST) network to restore the low-resolution compressed image, which jointly captures the hierarchical feature representations and enhances each-scale representation with Swin transformer, respectively. Moreover, we find that the pretraining with Super-resolution (SR) task is vital in compressed image super-resolution. To explore the effects of different SR pretraining, we take the commonly-used SR tasks (\egno, bicubic and different real super-resolution simulations) as our pretraining tasks, and reveal that SR plays an irreplaceable role in the compressed image super-resolution. With the cooperation of HST and pre-training, our HST achieves the fifth place in AIM 2022 challenge on the low-quality compressed image super-resolution track, with the PSNR of 23.51dB. Extensive experiments and ablation studies have validated the effectiveness of our proposed methods. The code and models are available at \url{https://github.com/USTC-IMCL/HST-for-Compressed-Image-SR}.
%Images from internet are often compressed with JPEG algorithm, and result in annoying block artifacts. To restore clean images, many strong image restoration networks are proposed and achieve state-of-the-art performance on compression artifacts reduction tasks. However, few of these studies considering handle compressed image super-resolution problem. To address this, we come up with HST, a hierarchical network using Swin Transformer, which has the ability to super-resolve a image while remove its compression artifacts. With the help of powerful Swin Transformer and the integration of information from different spatial scales, our proposed method achieves the fifth place in AIM 2022 challenge on low-quality compressed image super-resolution track, with PSNR of 23.75dB. Extensive experiments also demonstrate our method out perform existing image super-resolution networks.  

\end{abstract}

\section{Introduction}
Image super-resolution (SR) has achieved a quantum leap with the development of deep neural networks, which aims to restore the high-resolution (HR) images from their low-resolution counterparts. Existing SR can be roughly divided into three categories, simulated SR~\cite{SRCNN,EDSR,DRCN,RCAN,ESRGAN} (\egno, bicubic downsampling), real-world SR~\cite{RealSR,RealSR1,DrealSR,RealESRGAN,BSRGAN,li2021learningORNet,wei2020aim2020} and blind SR~\cite{KernelGAN,IKC,FKP,DASR,luo2022deep}, respectively. In particular, real-world and blind SR are greatly developed in recent years, of which the degradations are more consistent with unknown real-world distortions. However, not all images suffer from real-world degradation. In most cases, the images are susceptible to various compression artifacts together with low-resolution, since the image compression~\cite{wu2021learnedLBHIC,li2021taskTSC,pennebaker1992jpeg,rabbani2002overview,VVC}, transmission and storage. This hybrid degradation poses a challenging image process task, \ieno, compressed image super-resolution.
%Low-level computer vision tasks benefit a lot from the development of neural network techniques. One typical field is Singe image super-resolution (SISR), which aims to restore the high-resolution (HR) image from its low-resolution (LR) counterparts. Due to the requirement of high-resolution input for some downstream tasks, SISR has been incorporated into more and more real-world applications, such as medical image processing \cite{zhang2018fast}, surveillance image processing \cite{zhang2010super} and security image processing \cite{gohshi2015real}, etc. With the assistance of convolutional neural network (CNN), methods like (cite) SRCNN, EDSR, R... have demonstrated their great success in SISR task. Recently, transformer based methods (cite) SwinIR, IPT, Swin transformer, ViT... have achieved remarkable results in computer vision tasks, above all SwinIR \cite{liang2021swinir} is the state-of-the-art baseline on image restoration tasks, including SISR.

\begin{figure}[t]
    \centering
    \includegraphics[width=0.9\linewidth]{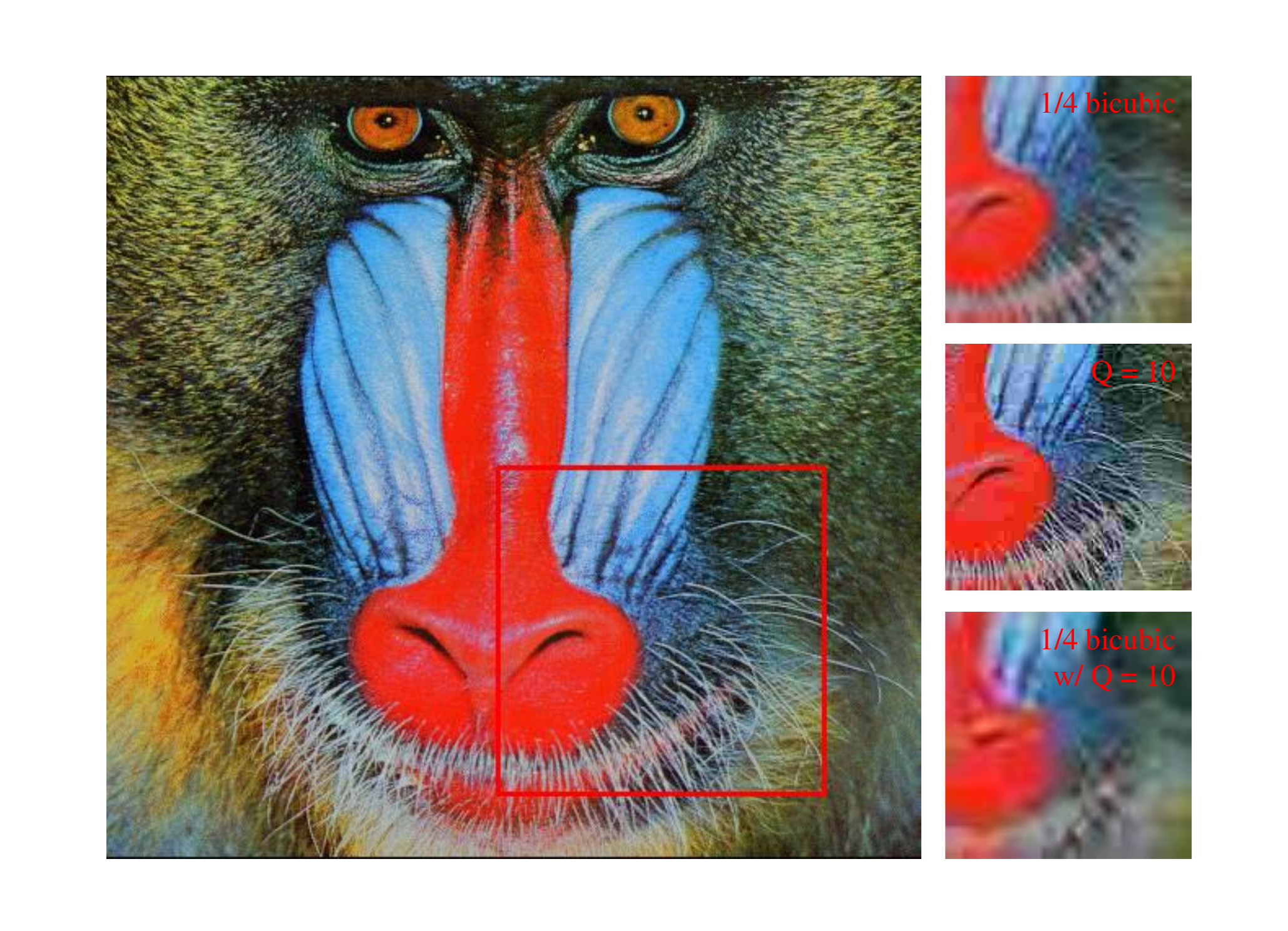}
    \caption{A comparison between different degradations. The left image is the high-resolution reference image. The images from the top right to bottom right are $1/4$ bicubic downsampling, JPEG compression with a quality factor of 10, and the combination of the above two distortions, respectively. Note that the bottom right image has the most severe degradation, thus requiring a stronger representation ability for the network to remove distortion.}
    \label{fig:intro}
\end{figure}

As shown in Fig.~\ref{fig:intro}, unlike general image SR and compression artifacts, the degradations of compressed low-resolution images are more severe, which composes of blurring, block artifacts, and noise, etc. Existing methods on image SR~\cite{EDSR,RCAN,SRCNN,ESRGAN} and compression artifacts removing~\cite{ARCNN,svoboda2016compression,cavigelli2017cas,DMCNN,DDCN} cannot work well on such brand-new degradation, since the large distribution shift. As the pioneering works, a series of works~\cite{zheng2022progressive,li2021comisr,wu2022animesr} began to investigate the compressed video super-resolution. To further promote the development of compressed image/video super-resolution, AIM2022~\cite{AIM2022} firstly holds the significant competition on compressed image super-resolution, where images are firstly down-sampled with the scale $1/4$, and then, are compressed with JPEG using an extreme low-quality parameter $Q=10$. A na\"ive and intuitive strategy to deal with it is exploiting a well-trained SR network and JPEG artifacts removing network to restore the distorted images in a sequential manner. However, the above strategy always fails since the distribution shift between hybrid distortions~\cite{li2020learning,liu2020lira} and single distortion.  Compressed image super-resolution requires that the restoration network have the strong  representation capability to learn structure and texture jointly.

%These methods usually have remarkable performance on well studied tasks, where input LR image has no or little JPEG compression artifact. However, in the practical scenarios, images are mostly downsampled and compressed by algorithms like JPEG \cite{pennebaker1992jpeg}, JPEG2000 \cite{rabbani2002overview} to save storage space on internet, thus the degradation contained in these images are more serious compared to that in the well-collected dataset. To reduce JPEG compression artifact, piles of works \cite{liu2019multi,cavigelli2017cas,ARCNN,DMCNN,gunawan2022cisrnet} have shown potential of using CNN, but few of them \cite{kang2013self,gunawan2022cisrnet,chen2018cisrdcnn} consider JPEG compression artifact removal with image super-resolution problem, let alone when these two degradations are severe (\egno compression quality $\mathcal{Q}=10$ combined with $\times4$ upsampling). State-of-the-art networks are mostly designed for JPEG compression artifact removal without change of the image spatial resolution, thus lack representation ability when directly applied to compressed image super-resolution task. Network should be designed to have more powerful feature extract ability to facing this puzzle, especially when degradation levels are severe, as shown in Fig. \ref{fig:intro}.

In this paper, we present Hierarchical Swin Transformer, namely HST, to tackle the compressed image super-resolution problem. 
Specifically, previous works~\cite{li2020multi,DMCNN} have shown superior advantages of hierarchical architecture on compression artifacts removing due to their great representation ability. Meanwhile, the variants of the transformer have been explored for image processing and quality assessment~\cite{chen2021IPT,liu2022swiniqa,lu2022rtn}, \egno, SwinIR~\cite{liang2021swinir}, which achieve remarkable performance compared with their CNN counterparts, since their capability of global contextual representation. Inspired by these, we present the Hierarchical Swin Transformer (HST) by incorporating the individual advantage of the above two architectures. In particular, 
%Inspired by previous works \cite{li2020multi,pang2020fan,liu2019multi,DMCNN}, which demonstrate that hierarchical network architecture has the ability of efficiently removing JPEG compression artifacts, we incorporate hierarchical network structure with recently published work \cite{liang2021swinir}, to form our HST. More specifically, 
our HST consists of four modules: hierarchical feature extraction module, feature enhancement module, fusion module, and HR reconstruction module. The hierarchical feature extraction module uses multiple convolution layers with different strides to obtain hierarchical feature maps at different scales. Then, the residual swin transformer block (RSTB) from SwinIR \cite{liang2021swinir} is used for the feature enhancement in each hierarchical branch. After getting the enhanced hierarchical features, we fuse them by concatenating the upsampling low-scale feature and high-scale feature, and then, input them into a convolution layer to obtain the fused feature.  Lastly, we can get the super-resolved HR image with the HR reconstruction module, which is composed of convolution layers and pixelshuffle layers. 

We also investigate the compressed image super-resolution from the perspective of pretraining. We observe that the pretraining with image super-resolution plays a vital role in the compressed image SR. Specifically, we systematically explore the effects of different image super-resolution tasks, including traditional SR, \ieno, bicubic downsampling, and two RealSR simulation methods from BSRGAN~\cite{BSRGAN} and DRTL~\cite{li2021few}. Extensive experiments reveal that the pretraining with the RealSR simulation from DRTL~\cite{li2021few} is better for compressed image SR.
%Hierarchical feature extraction uses multiple convolution layers with different strides to obtain feature maps at according scales. Then, residual swin transformer block (RSTB) from SwinIR \cite{liang2021swinir} is served as deep feature extractor for each of the hierarchical branch. Additionally, a residual skip connection is utilized to incorporate features extracted by first part, and features extracted by RSTB. Moreover, we use a convolution layer to integrate information from different hierarchical branches. Lastly, several convolution layers form as HR reconstruction part, to generate super-resolved images without compression artifacts.

The contributions of this paper can be summarized as:
\begin{enumerate}
\item We present the Hierarchical Swin Transformer (HST) for compressed image super-resolution, which incorporating the advantages of strong representation ability and global information utilization.
%\item We use swin transformer as our deep feature extractor with long-range dependency modelling, which is more suitable for severe compression artifacts removal and super-resolution.
\item We investigate compressed image super-resolution from the pretraining perspective. Based on the observation, we find one proper pretraining scheme for compressed image super-resolution.
\item Extensive experiment results show that our HST achieve a remarkable result on compressed image super-resolution task under heavy distortion (compression quality $Q=10$ combined with $1/4$ downsampling).
\end{enumerate}

\section{Related Works}
\subsection{Single Image Super-Resolution}
Single Image Super-resolution (SISR) has been developed expeditiously with the advances of deep neural networks. SRCNN~\cite{SRCNN}, as the pioneering work, firstly introduces the CNN to SISR and learns the network by minimizing the mean square error (MSE) between the generated images and their corresponding high-resolution (HR) images. Then, a series of works for SISR~\cite{EDSR,RCAN} are proposed by designing or modulating the network architecture.  EDSR~\cite{EDSR} revises the conventional residual module by removing the BatchNorm layers. RCAN~\cite{RCAN} adds channel attention to the residual blocks, which focus on more informative channels. And SAN~\cite{SAN} introduces the second-order channel attention to utilize the second-order feature statistics for more discriminative representations.

However, the above works exhibit poor capability for subjective quality improving. To tackle the above challenge, SRGAN~\cite{SRGAN} firstly introduces Generative Adversarial Network (GAN) to SISR, and adopts the adversarial loss for approximating the natural image manifold. As an improved version of SRGAN, ESRGAN~\cite{ESRGAN} exploits Relativistic average GAN~\cite{RAGAN} to enhance the discriminator and computes the VGG feature before the activation function to calculate the perceptual loss. To further improve the discriminator, FSMR~\cite{FSMR2022} comes up with feature statistics mixing regularization, which encourages the discriminator's prediction to remain invariant to the style of the input image.

Recently, real-world image super-resolution (RealSR)~\cite{RealSR,RealSR1,DrealSR,BSRGAN,RealESRGAN} and blind image super-resolution~\cite{IKC,FKP,KernelGAN,DASR} have been proposed to solve more severe and unknown hybrid distortions existed in real-world low-resolution images. To tackle unseen distortions (\ieno, blind distortions), KernelGAN~\cite{KernelGAN} train an internal-gan to estimate the degradation kernel contains in low-resolution images. IKC~\cite{IKC} ameliorates the estimation process into an iterative one, which can deal with more complex blind distortions. 
%To take a step further, FKP~\cite{FKP} leverages the degradation kernel priors supplement to network training, thus estimates unseen degradation kernels more accurately. 
Different from the aforementioned works that predict the degradation kernel, RealSR directly trains networks on synthesized real-world distorted image pairs, such as BSRGAN~\cite{BSRGAN} and ESRGAN~\cite{ESRGAN}.
%To achieve such goal, BSRGAN~\cite{BSRGAN} first to propose a practical degradation synthesize pipeline, which improve the network's robustness toward real-world distortions. Further on, real-ESRGAN~\cite{RealESRGAN} introduces a second-order degradation modeling for real distortion removal, and achieve remarkable result on benchmark datasets. 
In this paper, we focus on the compressed image super-resolution, which is more significant and valuable in the real world.

\subsection{Compression Artifacts Removal}
 Compression artifacts removal aims to remove the distortions caused by image/video codecs. Early works of compression artifact removal mainly focus on the design of manual filters in the DCT domain. Due to the success of CNN in image denoising and image super-resolution, Yu et al.~\cite{ARCNN} propose ARCNN, the first CNN-based method for compression artifacts removal. Svoboda et al.~\cite{svoboda2016compression} introduce residual learning to deepen the network under the assumption of ``deeper is better". However, ARCNN and its follow-up works only process artifacts in the pixel domain. DDCN~\cite{DDCN}, DMCNN~\cite{DMCNN} and ${D^3}$~\cite{wang2016d3} utilize DCT domain prior on the basis of pixel domain. Based on the network of extracting the dual domain knowledge, Fu et al.~\cite{fu2019jpeg} use dilated convolution for multi-scale feature extraction and added convolutional sparse coding to make the model more compact and explainable.  Recently, %Fu et al.~\cite{fu2021compression} introduce a variant model to formulate image de-blocking problem, leveraging two priors of the image content and gradient to remove compression artifacts. 
 there are a series of works that explore the hierarchical structures for compression artifacts removal. Lu et al.~\cite{lu2019learned} prove that adding multi-scale priors to the image restoration network can effectively eliminate compression artifacts. Inspired by Lu et al.~\cite{lu2019learned}, Li et.al add a non-local attention module to fuse multi-scale features effectively and obtain the post-processing network MSGDN~\cite{li2020multi} of VCC Intra coding. Based on the above excellent works, we also introduce the hierarchical module to our compressed image super-resolution network.

%However, when these above methods are applied to compression image super resolution, the mixed distortion of upsampling distortion and compression distortion is not considered, therefore they cannot be used directly.
\begin{comment}
    \subsection{Multi-scale Network}
The design of multi-scale networks has been widely applied in Vision tasks such as image segmentation~\cite{zhou2019unet++,tao2020hierarchical,he2019dynamic} and object detection~\cite{walambe2021multiscale,wang2018multiscale,gupta2021almnet}. Due to the ability of adaptively detecting image features at different scales, it has also been introduced into SISR~\cite{li2018multi}. Lu et al.~\cite{lu2019learned} proved that adding multi-scale priors to the image restoration network can effectively eliminate compression artifacts. By setting convolution kernels with different strides, feature maps of different scales can be obtained. After concatenating the feature of the second scale and the upsampled feature of the smallest scale, 
it can be fused by feeding into several residual blocks, and then fused with the feature of the third scale in the same way. Inspired by Lu et al.~\cite{lu2019learned}, Li et.al add non-local attention module to fuse multi-scale features effectively and obtain the post-processing network MSGDN~\cite{li2020multi} of VCC intra coding. Based on these works, we combine the multi-scale module into the compressed image super resolution network.
\end{comment}

\section{Method}
\begin{figure}[t]
    \centering
    \includegraphics[width=1.0\linewidth]{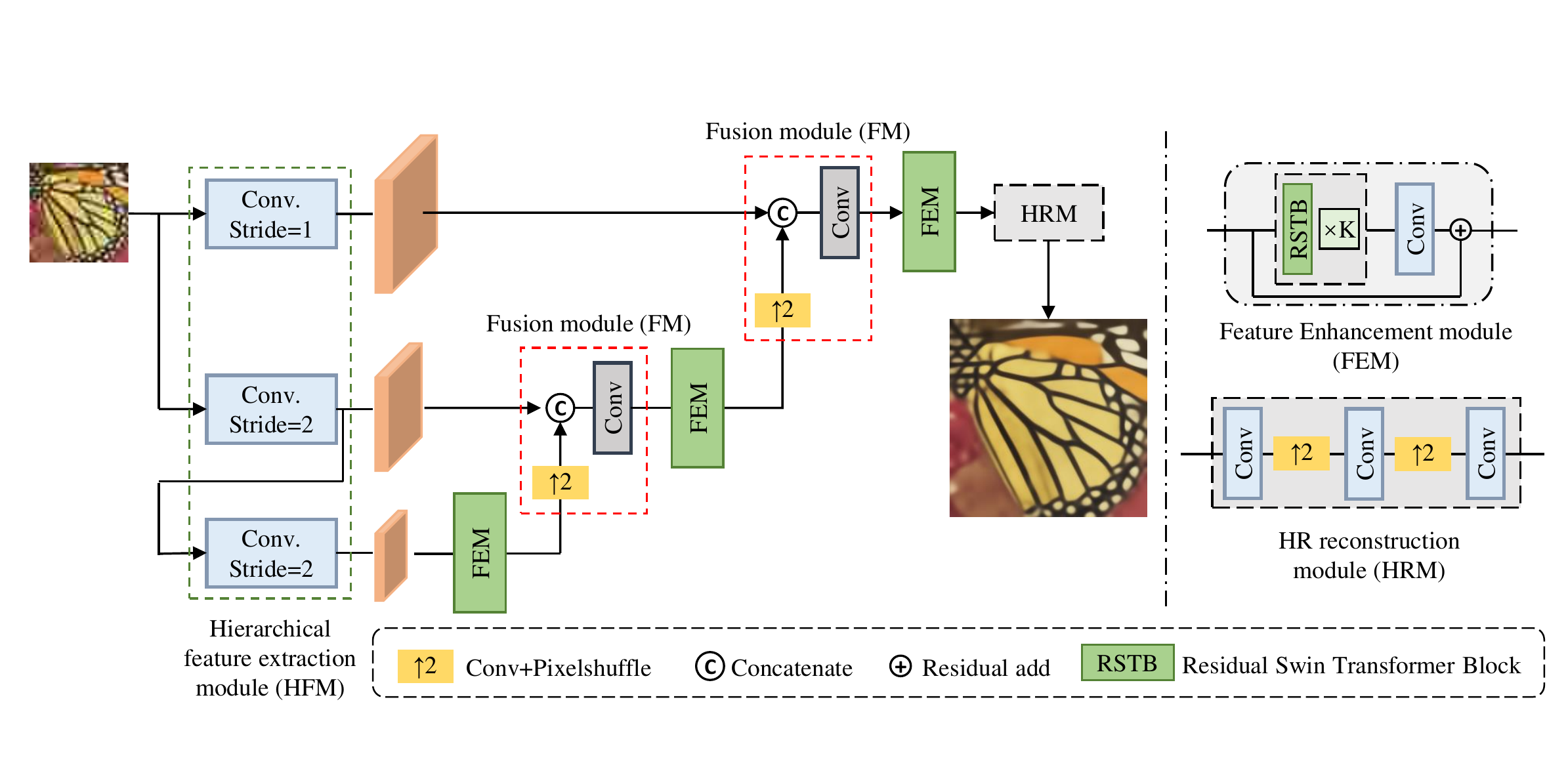}
    \caption{The architecture of proposed HST}
    \label{overall model}
\end{figure}

In this section, we will explain our HST and clarify our pretraining strategy in detail. As shown in Fig.~\ref{overall model}, our HST is composed of four main components, respectively as hierarchical feature extraction module (HFM), feature enhancement module (FEM), fusion module (FM) and HR reconstruction module (HRM).

%ur model can be summarized into three main components: hierarchical feature extraction, deep feature integration and HR reconstruction, as shown in Fig. \ref{overall model}.}

\subsection{Hierarchical Feature Extraction}

Previous works \cite{chen2019drop,pang2020fan} have revealed that extracting hierarchical features at different scales from images and processing them in a divide-and-conquer manner, can provide the network a strong representation ability. And thus, it can 
%extracting input LR image into multiple scales' features can benefit network for more powerful representation ability, thus able to 
deal with more severe and complex image degradation effectively. To achieve a trade-off between network parameters and performance, we choose a three-branches hierarchical architecture as our backbone. More specifically, the input LR image is gradually passed through three different convolution layers with different kernel sizes and strides, to extract the hierarchical representations with three scales. Following the implementation in \cite{pang2020fan}, we design upper branch convolution by $k7n60s1p3$, where $k, n, s, p$ stands for kernel size, number of channels, stride and padding respectively. For other two branches, we use convolution $k5n60s2p2$ to obtain the middle-scale feature map, and convolution $k3n60s2p1$ to obtain the low branch's feature map from the aforementioned feature map. The whole process can be formed as Eq. \ref{1}.

\begin{equation}\label{1}
\begin{aligned}
    F_{h} &= \mathrm{Conv}_{k7n60s1p3}(I_{l}) \\
    F_{m} &= \mathrm{Conv}_{k5n60s2p2}(I_{l}) \\    
    F_{l} &= \mathrm{Conv}_{k3n60s2p1}(F_{m})\\
\end{aligned}
\end{equation}
where $I_{l}  \in \mathbb{R}^{H \times W \times C}$ is the compressed low resolution image and $H$, $W$, $C$ refer to its height, width and color channel, respectively.  $F_{h}$, $F_{m}$, $F_{l}$ represent the features of three branches.

Through this process, we obtain the hierarchical features $\{F_{h}, F_{m}, F_{l}\}$ at different scales. Then, we will input them into the feature enhancement module to process in a divide-and-conquer manner.

\subsection{Feature Enhancement and Fusion}\label{Fi}
The feature enhancement module and feature fusion module are the important components of HST. We will clarify them carefully in this section.
\subsubsection{Feature enhancement module} 
Different from previous hierarchical networks \cite{liu2019multi,li2020multi,DMCNN,pang2020fan} for image restoration and super-resolution, where convolutional neural network (CNN) is used as the feature enhancement module for each branch, we use swin transformer architecture as ours. As proved by \cite{liang2021swinir}, swin transformer-based architecture can model long-range dependency enabled by the shifted window mechanism. Therefore, this architecture is more suitable for difficult degradation removal tasks, \egno, compressed image super-resolution. Moreover, with the help of swin transformer, we can get better performance with less parameters.
\begin{figure}[ht]
    \centering
    \includegraphics[width=1.0\linewidth]{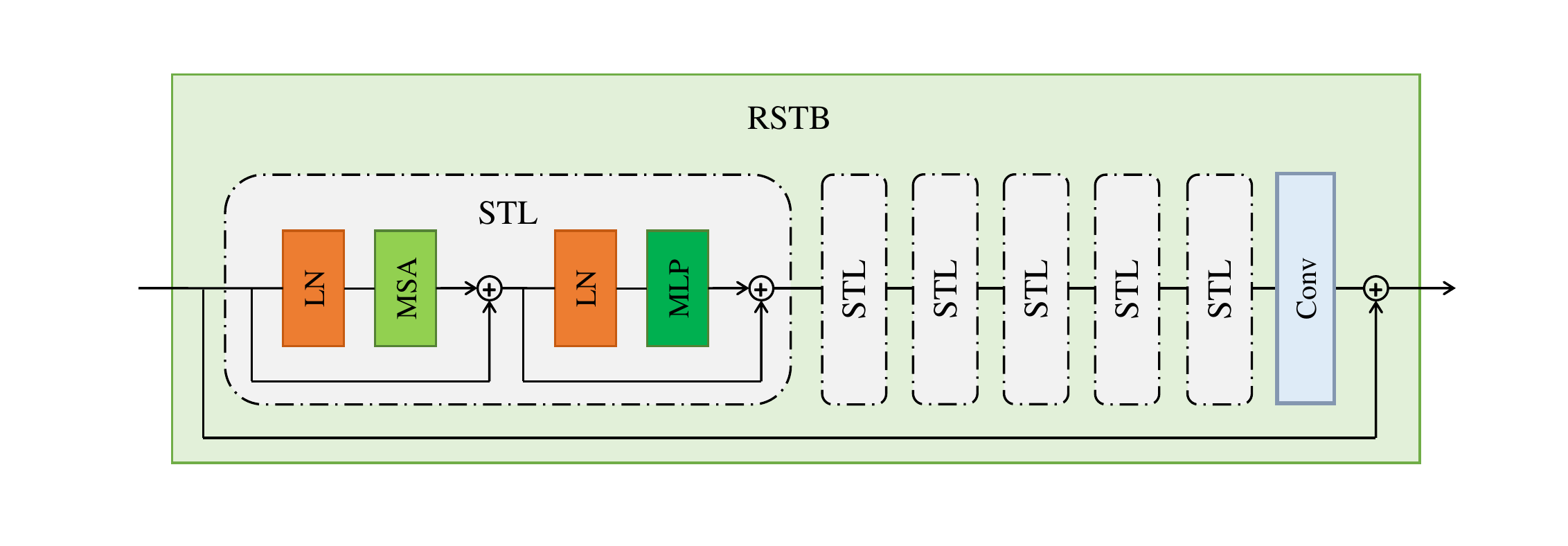}
    \caption{Network structure of residual swin transformer block (RSTB)}
    \label{RSTB}
\end{figure}

Specifically, we directly apply multiple residual swin transformer blocks (RSTB) from \cite{liang2021swinir}, as our feature enhancement module. The architecture of RSTB is shown in Fig. \ref{RSTB}. Each RSTB is composed of several swin transformer layers (STL), a convolution layer and a  residual skip connection. This process can be formulated as Eq. \ref{2}

\begin{equation}\label{2}
\centering
\begin{aligned}
   & F_0  = F_{in} \\
   & F_i  = \mathrm{STL}(F_{i-1}), \quad  i = 1,2, \dots, K \\
   & F_{out}  = \mathrm{Conv}(F_K) + F_0 \\ 
\end{aligned}
\end{equation}
where $F_{in}$ is the input feature of one STL layer, and $\mathrm{STL}(\cdot)$ means each STL layer inside RSTB, which can be formulated as Eq. \ref{3}

\begin{equation}\label{3}
\begin{aligned}
   X = \mathrm{MSA}(\mathrm{LN}(X)) \\
   X = \mathrm{MLP}(\mathrm{LN}(X)) \\ 
\end{aligned}
\end{equation}
where $\mathrm{MSA}(\cdot)$ stands for multi-head self-attention, $\mathrm{MLP}(\cdot)$ stands for a multi-layer perceptron with two fully-connected layers and GELU as activation, and $\mathrm{LN}(\cdot)$ stands for LayerNorm. Since STL is not our contribution and previous works have already proved the effectiveness of this module, we directly utilize the same architecture as is presented in \cite{liang2021swinir}.

\subsubsection{Feature fusion module}
After getting the enhanced features $F_{l}^*$ from the low-branch enhancement module, composed of severe RSTB blocks. We will integrate it into the higher feature with the fusion module, which aims to bring the contextual information from low-scale to high-scale. To demonstrate the fusion process clearly, we take the fusion of the low-branch feature and the middle-branch feature as an example. As described in Eq.~\ref{4}, the low-branch feature $F_l$ is enhanced to $F_{l}^{*}$ with low-branch feature enhancement module $FEM_l$. Then we concatenate the super-resolved low-branch feature $F^{*}_{l\ \uparrow 2}$ and middle-branch feature $F_m$, and exploit the convolution layer to fuse these two components. Finally, we can obtain the enhanced middle-branch feature $F_m^*$ by passing the fused feature $F_m$ into the middle-branch enhancement module $FEM_m$. It is worthy to notice that, the up-sampling operation is implemented with Pixelshuffle~\cite{shi2016pixelshuffle}, which can bring more stable results. The fusion of middle branch and high-branch features are implemented in the same way in Eq.~\ref{2}.

%After we extract deep feature by several RSTBs of certain branch, we integrate this feature map with upper branch, to fusion information from different spatial resolutions, thus improve the representation ability of our model. To be specific, 
%we take this integration process between Minimum branch and Middle branch as an example, as shown in Eq. \ref{4}

\begin{equation}   \label{4}
\begin{aligned}
   F_{l}^* & = \mathrm{FEM_l} (F_{l}),  \\
   F_{m} & = \mathrm{Conv} (F_{m}\  \textcircled{c}\  F^*_{l\ \uparrow2})\\ 
   F_{m}^* &= \mathrm{FEM_m} (F_{m}) \\
\end{aligned}
\end{equation}
%Here, first equation demonstrate the deep feature extraction process conduct on $F_{\mathrm{low}}$. In order to integrate features between different branches, an upsample operation is performed on lower branch's feature map, and then concatenate with upper branch's feature map channel wise. The upsample operation we used here is PixelShuffle \cite{shi2016pixelshuffle}, which can provide a more stable upsample result. Notice that, the upper branch's feature here (\ieno, $F_{\mathrm{middle}}$) are not yet passed through deep feature extractor. This operation can better integrate information from different branches by a coarse-to-fine manner. Lastly, we use a convolution layer to fuse these two feature maps, which enable information interaction, as shown in third line of Eq. \ref{4}.

\subsection{HR Reconstruction Module}
Since the compressed image super-resolution in the competition requires the resolution of network output to be $4\times$ higher than their input image, the HR reconstruction module aims to produce the final three-channel RGB high-resolution clean image with the enhanced high-branch feature $F_h^*$.

%As its name demonstrates, this part is utilized to reconstruct high-resolution output image according to network's restored feature map. In HST, this part is basically target for super-resolving features incorporated by deep feature integration part, and output a three-channel RGB high-resolution artifacts-free image.

As shown in Fig. \ref{overall model}, this part is composed of two sub-pixel convolution layers, including two convolution layers and two PixelShuffle layers~\cite{shi2016pixelshuffle}. Following the previous SR works~\cite{EDSR,RCAN,liang2021swinir}, we utilize two sub-pixel convolution layers for the $4\times$ upsampling. Finally, a convolution layer is used to generate the output HR image.

\subsection{Pretraining with SR}\label{3.4}
%Pretraining, as a simple but effective strategy, has been applied to image processing~\cite{}. Inspired by this, 
To further boost the capability of the network, We also explore one simple but effective pertaining strategy for compressed image super-resolution. It is noteworthy that pretraining with more relevant distortions can bring better knowledge transfer. Particularly, we select three SR tasks as the pretraining schemes, \ieno, traditional SR, two RealSR simulation methods from BSRGAN~\cite{BSRGAN} and DRTL~\cite{li2021few}, and explore their effectiveness for compressed image super-resolution. The relevant experimental analyses are shown in Sec.~\ref{pretrain}, which demonstrates pretraining with RealSR simulations leads to promising results, especially with the simulation in DRTL~\cite{li2021few}. 
%Since directly training the network on such complex distortion scenario is difficult, which may cause a non-convergence problem during optimization, we utilize a pretraining-finetuning scheme. To obtain a good start-point for compressed image super-resolution, we first pretrain HST on simpler super-resolution tasks (\ieno, bicubic image SR), and then finetune the network based on pretrained weight. More details are given in Sec. \ref{pretrain}

%Specifically, following the design in work \cite{liang2021swinir}, we first use a convolution layer to reduce channel number before upsampling, and then exploit PixelShuffle \cite{shi2016pixelshuffle} operation  twice for our $\times4$ compressed image super-resolution task. Finally, a convolution layer is used to generate output HR image.

\subsection{Loss Functions}
In order to enable our HST to be competent for the task of compressed image super-resolution, we first pretrain our network on the $\times4$ super-resolution task, and then finetune it for compressed image super-resolution. 
For the $\times4$ super-resolution pretraining, we optimize network parameters by minimizing the $L_1$ pixel loss as:

\begin{equation}
    \mathcal{L} = \| I_{SR} - I_{HR}\|_{1},
\end{equation}
where $I_{SR}$ is obtained by passing low-resolution images through the network, and $I_{HR}$ is the corresponding ground-truth HR image. For compressed image super-resolution, we optimize network parameters by minimizing the Charbonnier loss \cite{charbonnier1994two}.

\begin{equation}
    \mathcal{L}=\sqrt{\left\|I_{SR}-I_{HR}\right\|^{2}+\epsilon},
\end{equation}
where $\epsilon$ is set as default value $10^{-9}$.

\section{Experiments}
\subsection{Datasets}\label{Data}
We produce the experimental results in our paper with two training datasets, DIV2K~\cite{DIV2K} (including 800 high-resolution images) and Flick2K~\cite{Flickr2K} (including 2650 high-resolution images).
%We train our network with a combination of 800 high-resolution images from DIV2K \cite{DIV2K}, and 2650 high-resolution images from Flickr2K \cite{Flickr2K}.
In the competition AIM2022~\cite{AIM2022}, we also collect extra 746 high-resolution images from CLIC 2021 official website\footnote{http://clic.compression.cc/2021/tasks/index.html} as the additional training data, which is only used for the competition results in Sec.~\ref{AIM}. 
%Ours HST with whole training data is shown in Sec.~\ref{AIM}.
%To further improve the performance and compare with the single branch network we used in AIM2022~\cite{AIM2022} challenge, we collect training and validation dataset from CLIC 2021 official website\footnote{http://clic.compression.cc/2021/tasks/index.html}, a total of 746 high-resolution images to form an additional training dataset. 
%%This additional dataset is only used for comparison with our method proposed during competition in Sec. \ref{AIM}.
For the testing stage in this paper, we adopt Set5~\cite{Set5}, Set14~\cite{Set14}, BSD100~\cite{BSD100}, Urban100~\cite{urban100}, Manga109~\cite{manga109} and DIV2K~\cite{DIV2K} validation as our testing datasets. 
%We include DIV2K validation dataset in the testing since it is used for performance evaluation in AIM2022~\cite{AIM2022} competition. Notice that, the height and width of some images from Manga109 dataset are not the multiple of 4, we crop these images to avoid possible misalignment between SR and HR.

\subsection{Implementation Details} 
We use a three-branch HST for our experiments. The channel numbers of three feature enhancement modules $FEM_{h},\  FEM_{m},\  FEM_{l}$ are set to 60, 60, 60, respectively. The spatial resolution of the high branch is 64$\times$64, and halved for each downscale branch. Following \cite{liang2021swinir}, we set the number of swin transformer layers (STLs) as 6 for all residual swin transformer blocks (RSTBs) in HST. We use 2, 4, and 6 RSTBs for low branch, middle branch and high branch, respectively. The window size is set to 8 throughout the experiment.

We train our HST using four NVIDIA 2080Ti GPUs, with a batch size of 16. We offline generate training image pairs by the MATLAB bicubic kernel, then add JPEG compression with specified quality factor through the OpenCV function. We randomly crop LR into 64$\times$64 patches for training. For data augmentation, we leverage random  flipping and random rotation simultaneously. In the stage of pretraining, the total training iterations are set to 400K. We adopt Adam optimizer with $\beta_1 = 0.9$ and $\beta_2 = 0.999$, the initial learning rate is set to 2e-4 and reduced by half at [100K, 250K]. In the stage of finetuning, we load network parameters from the pretraining stage. We conduct experiments on four different compression levels, with quality factors at 40, 30, 20 and 10, respectively. The training is first finished on quality factor at 40, with the initial learning rate and total iterations as 1e-4 and 200K. And the learning rate is halved after 100K iterations. The rest of tasks are finetuned based on the first task (\ieno, quality factor at 40), with the learning rate as 8e-5 and total iterations as 100K.

\subsection{Effects of different pretraining schemes}\label{pretrain}

\begin{table}[t]
\centering
\caption{Quantitative comparison for ablation study of network pretraining scheme. Results are tested on $\times4$ with compression quality 10 on Urban100~\cite{urban100} dataset in terms of PSNR/SSIM. Best performance  are in \tcl{red}.}
\setlength{\tabcolsep}{2.0mm}
\begin{tabular}{c|cccc}
\toprule
\multirow{2}{*}{Task} &  \multicolumn{4}{c}{Methods(PSNR/SSIM)} \\ \cline{2-5}  & w/o & bicubic $\times4$ & BSRGAN~\cite{BSRGAN} & DRTL~\cite{li2021few} \\ 
\midrule
$\times4$, Q=10  &   19.70/0.5181  &  19.92/0.5301 &  20.04/0.5375  &  \tcl{20.06/0.5383} \\
\bottomrule
\end{tabular}
\label{quantitative_pre}
\end{table}

\begin{figure}[h!]
    \centering
    \includegraphics[width=1.0\linewidth]{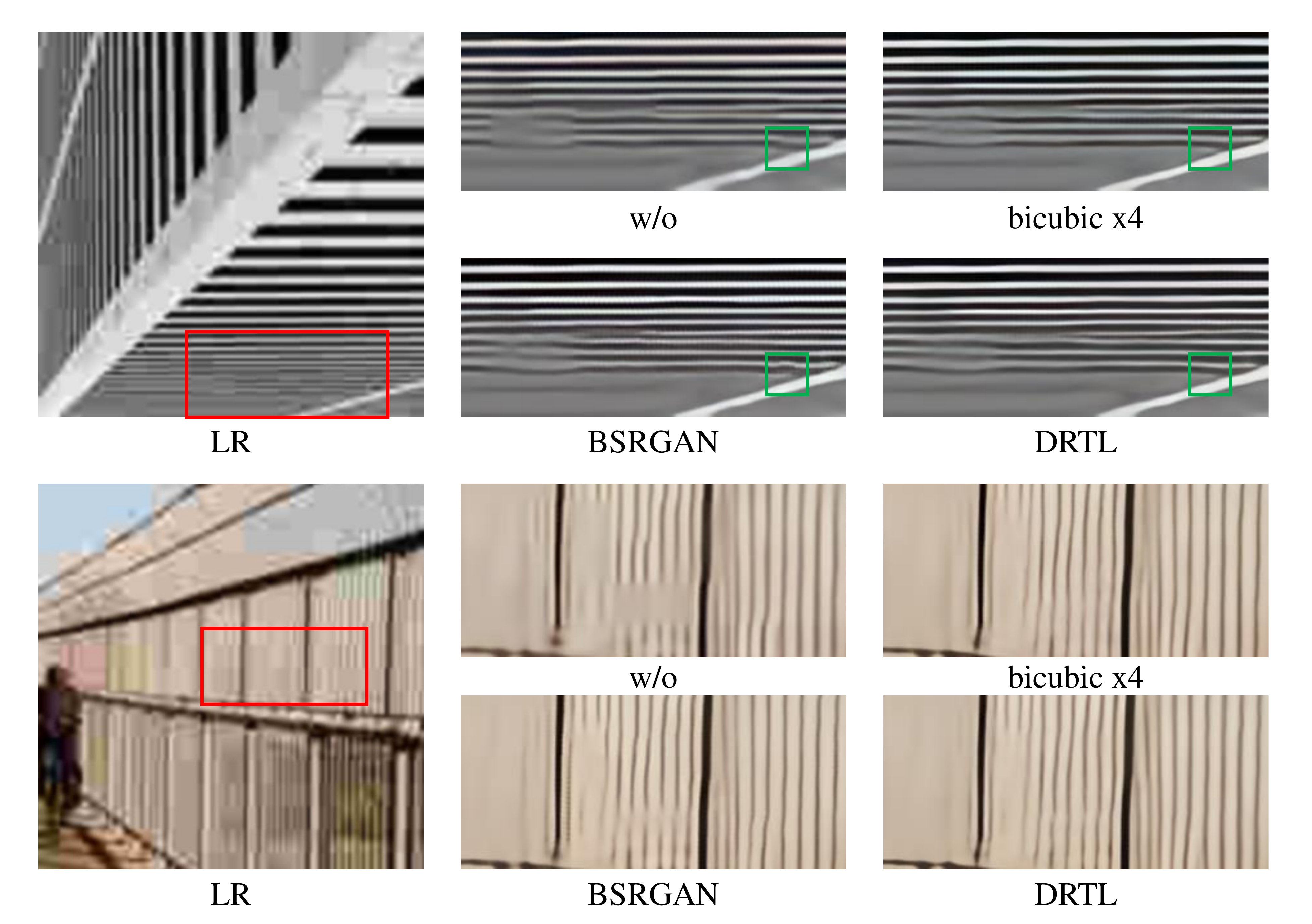}
    \caption{Qualitative comparison for different pretraining schemes on $\times4$ image super-resolution with compression quality 10. Testing images are ``011" and ``024" from Urban100~\cite{urban100} respectively. }
    \label{ablation_pre_img}
\end{figure}

As discussed in Sec. \ref{3.4}, pretraining is crucial for compressed image super-resolution task. To find out the optimal pretraining scheme, we conduct an ablation study on four different strategies, including: without pretraining, pure bicubic $\times4$ pretraining, pretraining with RealSR simulation from BSRGAN~\cite{BSRGAN}, and pretraining with RealSR simulation from DRTL~\cite{li2021few}. BSRGAN~\cite{BSRGAN} uses a practical complex degradation simulation process, which demonstrates its effectiveness on real-world distortion removal. DRTL~\cite{li2021few} proposes a multi-task degradation training scheme, to simulate distortion in real-world scenarios, and works well on few-shot real-world image super-resolution problems. For fast convergence and convincing results, we use SwinIR-s~\cite{liang2021swinir} as a training model, and test network performance on Urban100~\cite{urban100}.

As shown in Table \ref{quantitative_pre}, quantitative results show that the pretraining with RealSR leads to a gain of 0.36dB/0.0202 on the test dataset, which reveals that the pretraining is vital for compressed image super-resolution. 
Another observation is that pretraining with RealSR can achieve a better performance compared with simple bicubic downsampling, especially with the simulation in DRTL~\cite{li2021few}. The reason for this might be that RealSR simulations contain lots of hybrid distortions, which are more complex and the knowledge is more likely to be transferred to the severely compressed image super-resolution task. 

%As expected, without pretraining, network suffers from non-convergence problem, results in poor performance. 
%Pretraining with only bicubic $\times4$ task 
%causes network to overfit on simple super-resolution task, thus can not deal with compression artifacts properly. Moreover, as shown in Fig. \ref{ablation_pre_img}, without pretraining, there are still block artifacts in the SR. Although BSRGAN~\cite{BSRGAN} and DRTL~\cite{li2021few} achieves similar quantitative results on test dataset, we can observe that BSRGAN~\cite{BSRGAN} generates broken lines and punctate artifacts aside some edges. This is because network pretrained with BSRGAN~\cite{BSRGAN} may treat serious compression artifacts as other types of distortions, and generate incorrect SR. On the contrary, DRTL~\cite{li2021few} learns the invariant information between multiple pretraining tasks, thus can fast adapt to compressed image super-resolution problem, generate more natural-looking images.

\subsection{Effects of hierarchical architecture}

\begin{figure}[t!]
    \centering
    \includegraphics[width=1.0\linewidth]{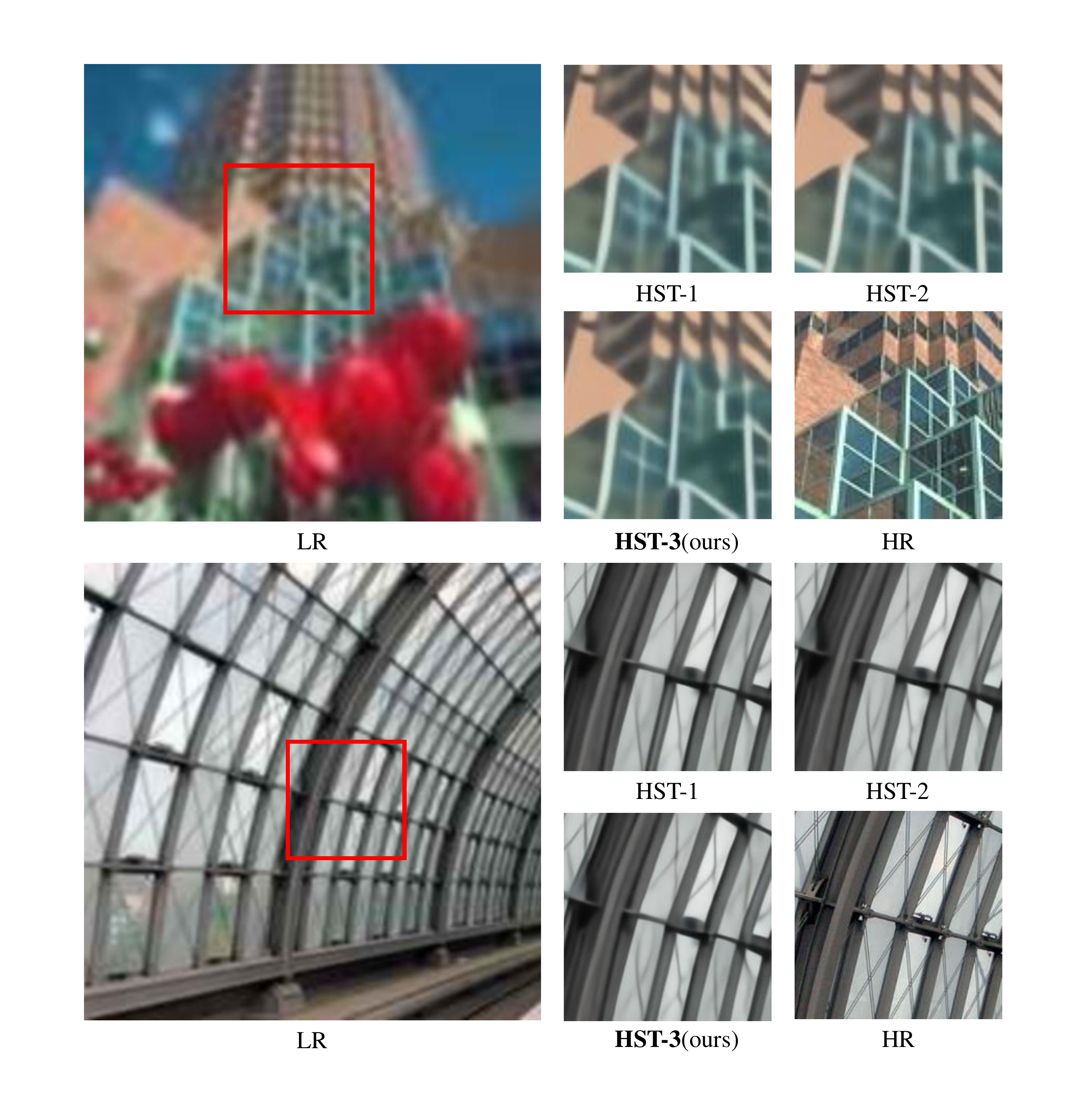}
    \caption{Qualitative comparison for different network scales on $\times4$ image super-resolution with compression quality 40. Testing images are ``095" from BSD100~\cite{BSD100} and ``002" from Urban100~\cite{urban100} respectively. }
    \label{ablation_img}
\end{figure}

To explore the advantage of introducing a hierarchical network structure, we set network branches from 1 to 3 and observe their performances on $\times4$ super-resolution with compression quality 40. Note that, one branch framework is almost the same as SwinIR-M~\cite{liang2021swinir}. As shown in Table \ref{quantitative_hierarchical}, more branches lead to higher performance. However, it also brings an increase of computational complexity. In this paper, we choose a three-branch HST to achieve the best performance. In addition, benefits from structure and texture information compensation from lower branches, three-branch HST can generate images with clearer lines and more structural components, as shown in Fig. \ref{ablation_img}

%benefits from the increase of branch number. However, when keep adding new branches into network structure, the window size of swin transformer will become larger than the input image's resolution, which may cause feature extractor to capture meaningless information, undermining feature fusion process. Therefore, we choose a three-branch HST to achieve best performance. In addition, benefits from structure and texture information compensation from lower branches, three-branch HST can generate images with clearer lines and more structural components, as shown in Fig. \ref{ablation_img}

\begin{table}[h]
\centering
\caption{Quantitative comparison for ablation study of network scales. The number of parameters is listed in the bracket. Results are tested on $\times4$ with compression quality 40 in terms of PSNR/SSIM. Best performance  are in \tcl{red}.}
\setlength{\tabcolsep}{1.1mm}
\begin{tabular}{c|c|ccccc}
\toprule
\multirow{2}{*}{Methods} & \multirow{2}{*}{Q} & \multicolumn{5}{c}{Datasets(PSNR/SSIM)} \\ \cline{3-7}  &   & Set5 & Set14 & BSD100 & Urban100 & Manga109\\ 
\midrule
HST-1(11.90M) & \multirow{3}{*}{\textbf{40}} &   25.28/0.726  &  23.78/0.613&   23.82/0.583 &   22.21/\tcl{0.652} &  23.69/0.767 \\
HST-2(12.98M)   &  &  25.35/0.727  & 23.82/\tcl{0.614}  & 23.84/\tcl{0.584}  & 22.21/0.651       & 23.78/\tcl{0.769}\\
HST-3(16.58M) &  & \tcl{25.39}/\tcl{0.728}  &  \tcl{23.84}/\tcl{0.614}   &   \tcl{23.87}/\tcl{0.584}  & \tcl{22.23}/0.651 &  \tcl{23.85}/0.768 \\
\bottomrule
\end{tabular}
\label{quantitative_hierarchical}
\end{table}

\subsection{Comparison with other frameworks}

\begin{figure}[t!]
    \centering
    \includegraphics[width=1.0\linewidth]{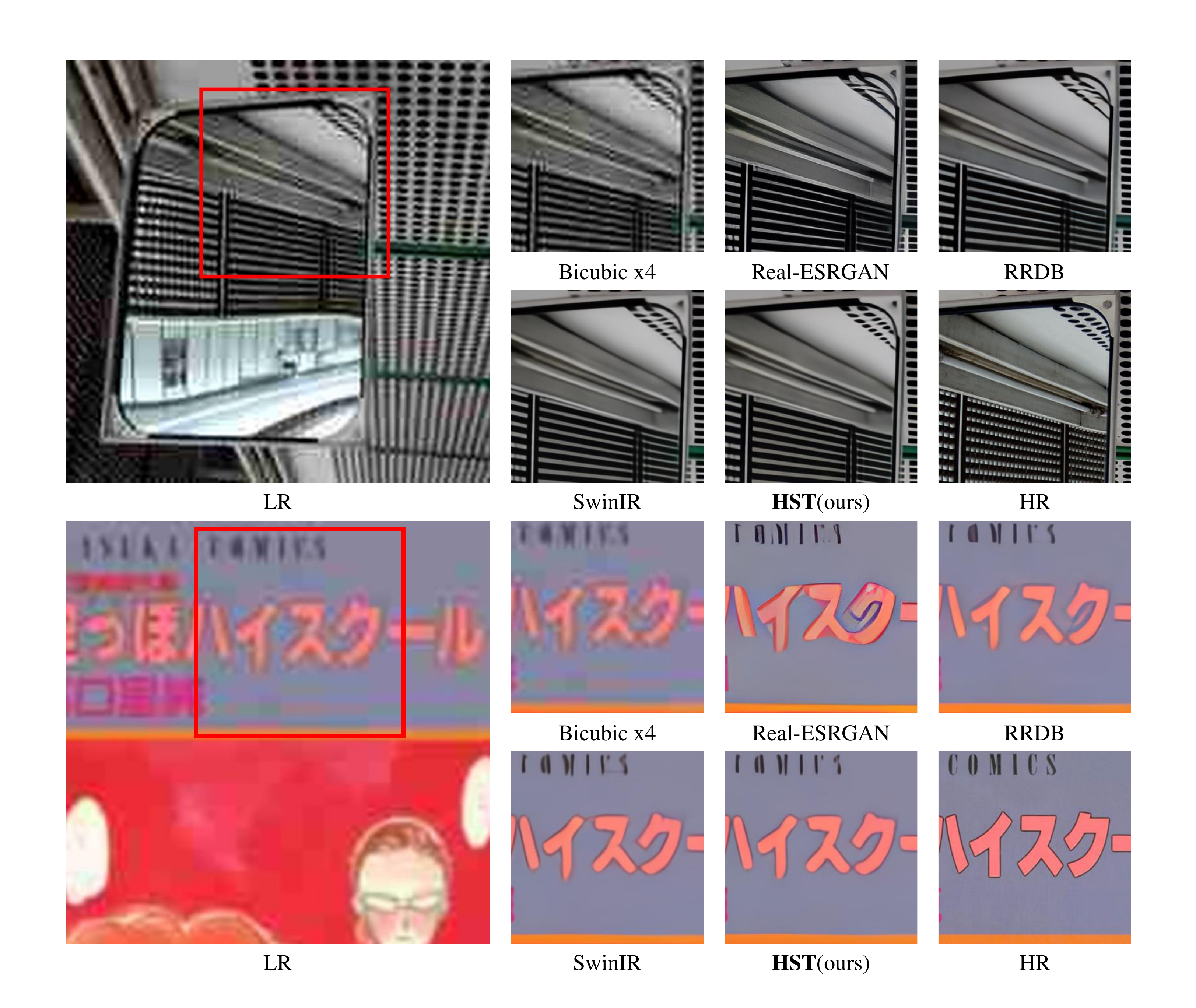}
    \caption{Qualitative comparison for $\times4$ image super-resolution with compression quality 20. Testing images are ``004" from Urban100~\cite{urban100} and ``KarappoHighschool" from Manga109~\cite{manga109}, respectively.  }
    \label{qualitative_img}
\end{figure}

We compare our HST with two other state-of-the-art models in image super-resolution, and one real-SR method~\cite{RealESRGAN} for qualitative comparison. Among them, RRDB~\cite{ESRGAN} uses residual in residual dense blocks to deepen the network structure, thus having the ability to better aggregate image structure and texture information from multi-levels. SwinIR~\cite{liang2021swinir} introduces transformer into image restoration tasks and outperforms previous CNN-based models. The performances are tested on Set5~\cite{Set5}, Set14~\cite{Set14}, BSD100~\cite{BSD100}, urban100~\cite{urban100}, Manga109~\cite{manga109}, respectively, with PSNR and SSIM in RGB channels. Moreover, we also test three models' performance on AIM2022~\cite{AIM2022} official validation dataset, which includes 100 images from the DIV2K validation dataset. Quantitative and qualitative results are shown in Table \ref{quantitative},\ref{quantitative_div2k} and Fig. \ref{qualitative_img}, respectively. We denote the model using a self-ensemble strategy~\cite{EDSR} with $^*$.

Extensive experiments show that our HST outperforms other methods by 0.25dB at most on compressed image super-resolution tasks. Even without self-ensemble, HST can still achieve an increase of 0.16dB at most. As shown in Fig. \ref{qualitative_img}, Real-ESRGAN~\cite{RealESRGAN} generates unnatural textures although its degradation process includes JPEG compression. Compared with other methods, our HST can generate SR with fewer artifacts. Moreover, HST performs better in rich texture areas, resulting in pleasant perception. All these benefit from a hierarchical network structure, which captures features at different scales and enhances the network's representation ability.

\begin{table}[t!]
\centering

\caption{Quantitative comparison for compressed image super-resolution on benchmark datasets. Results are tested on $\times4$ with different compression qualities in terms of  PSNR/SSIM. Best performance are in \tcl{red}.}
\setlength{\tabcolsep}{1.5mm}
\begin{tabular}{l|c|ccccc}
\toprule
\multirow{2}{*}{Methods} & \multirow{2}{*}{Q} & \multicolumn{5}{c}{Datasets(PSNR/SSIM)} \\ \cline{3-7}  &   & Set5 & Set14 & BSD100 & Urban100 & Manga109\\ 
\midrule
RRDB~\cite{ESRGAN} & \multirow{4}{*}{\textbf{10}} &   22.36/0.629         &     21.75/0.538           &       22.13/0.514           & 20.24/0.553        & 20.66/0.677 \\
SwinIR~\cite{liang2021swinir}   &         &     22.45/0.636         &     21.79/0.541        &         22.16/0.517         & 20.35/\tcl{0.561}       & 20.81/0.685           \\
\textbf{HST}     &   & 22.49/\tcl{0.637}     & 21.84/\tcl{0.542} & 22.18/0.517       & 20.38/0.559        & 20.88/0.684          \\
\textbf{HST$^*$} &    & \tcl{22.51/0.637} & \tcl{21.86/0.542} & \tcl{22.20/0.518} & \tcl{20.43/0.561}        & \tcl{20.94/0.686}             \\ \midrule \midrule

RRDB~\cite{ESRGAN} & \multirow{4}{*}{\textbf{20}} &  23.73/0.674  &   22.81/0.575 &  23.06/0.550  &      21.17/0.599            &   22.17/0.722  \\
SwinIR~\cite{liang2021swinir}    &    &      23.81/0.682 &         22.87/0.577       &  23.09/\tcl{0.551}   &    21.32/\tcl{0.608}  &          22.35/\tcl{0.729} \\
\textbf{HST} &  & 23.91/0.683 &   22.93/0.578 & 23.11/\tcl{0.551} & 21.33/0.607 &   22.41/0.728  \\
\textbf{HST$^*$}  &  &   \tcl{23.96/0.684}        &     \tcl{22.95/0.579}           &   \tcl{23.13/0.551}   &   \tcl{21.38}/0.607 &     \tcl{22.48/0.729}                 \\ \midrule \midrule
RRDB~\cite{ESRGAN} & \multirow{4}{*}{\textbf{30}}      &       24.74/0.708        &    23.42/0.599            &     23.53/0.569             &     21.77/0.630  &  23.09/0.750   \\
SwinIR~\cite{liang2021swinir}  &  & 24.83/0.713 &  23.43/0.600  &   23.53/\tcl{0.571}  &  21.85/\tcl{0.636}    &     23.20/0.755                 \\
\textbf{HST} &  & 24.89/0.713 &  23.49/0.600  &  23.57/\tcl{0.571}   &  21.91/0.635          &  23.30/0.754  \\
\textbf{HST$^*$} &   &    \tcl{24.94/0.714}           &   \tcl{23.52/0.601} &  \tcl{23.59/0.571}  &   \tcl{21.96/0.636} & \tcl{23.39/0.756}  \\ \midrule \midrule
RRDB~\cite{ESRGAN} & \multirow{4}{*}{\textbf{40}}      &    25.05/0.717           &      23.67/0.609          &    23.78/0.581             &     21.93/0.638             &23.37/0.756   \\
SwinIR~\cite{liang2021swinir}  &  & 25.28/0.726  &  23.78/0.613&   23.82/0.583 &   22.21/0.652 &  23.69/0.767 \\
\textbf{HST} &  & 25.39/0.728  &  23.84/\tcl{0.614}   &   23.87/0.584  & 22.23/0.651 &  23.85/0.768  \\
\textbf{HST$^*$}  &  &  \tcl{25.43/0.729}    &  \tcl{23.87/0.614}  &   \tcl{23.89/0.585}  &  \tcl{22.29/0.653} &  \tcl{23.94/0.771}  \\ \bottomrule
\end{tabular}
\label{quantitative}
\end{table}

\begin{table}[h!]
\centering
\caption{Quantitative comparison for compressed image super-resolution on DIV2K~\cite{DIV2K} validation datasets. Results are tested on $\times4$ with different compression qualities in terms of  PSNR/SSIM. Best performance are in \tcl{red}.}
\setlength{\tabcolsep}{1.5mm}
\begin{tabular}{c|c|cccc}

\toprule
\multirow{2}{*}{Datasets} & \multirow{2}{*}{Q} & \multicolumn{4}{c}{Methods(PSNR/SSIM)} \\ \cline{3-6}  &   & RRDB & SwinIR & \textbf{HST} & \textbf{HST$^*$}\\ 
\midrule
\multirow{4}{*}{DIV2K~\cite{DIV2K}} & \textbf{10} &   23.52/0.6400         &     23.57/0.6436           &       23.62/0.6436           & \tcl{23.65/0.6443}  \\
  &    \textbf{20}     &    24.68/0.6746  &     24.73/0.6771        &   24.77/0.6769       & \tcl{24.80/0.6777} \\
 &  \textbf{30} & 25.31/0.6949    & 25.32/0.6966 & 25.38/0.6963       & \tcl{25.41/0.6971} \\
 &  \textbf{40} & 25.58/0.7038 & 25.67/0.7077 & 25.74/0.7085 & \tcl{25.78/0.7093} \\ \bottomrule
\end{tabular}
\label{quantitative_div2k}
\end{table}

\subsection{AIM2022 challenge}\label{AIM}

To further explore the performance of our HST, we follow the training process we used in AIM2022~\cite{AIM2022} competition to train our HST. More specifically, we use all three training datasets described in Sec. \ref{Data} to finetune the network. After 100k iterations' finetuning with Charbonnier Loss~\cite{charbonnier1994two}, we further use MSE Loss to optimize the network until convergence. The result on the official validation dataset shows that, with hierarchical network architecture, HST outperforms the one-branch network we used in the competition by 0.05dB, with a final PSNR of 23.80dB.

\section{Conclusion}
In this paper, we propose the Hierarchical Swin Transformer for compressed image super-resolution, which incorporates the advantages of the hierarchical structure and Swin Transformer. Moreover, we find that pretraining with SR is vital and effective for compressed image super-resolution. Particularly, we explore three pretraining tasks, \ieno, traditional SR, and two RealSR simulations from BSRGAN and DRTL, respectively, of which the experimental results show that pretraining with RealSR simulations can bring better performance, especially with the simulation in DRTL~\cite{li2021few}. Extensive experiments demonstrate that, with a pretraining and hierarchical network structure, our HST achieves the best performance on compressed image super-resolution tasks. In addition, our model achieves the fifth place in the AIM2022 challenge, with a PSNR of 23.51dB.

\begin{comment}
\section*{Acknowledgement}
Acknowledgement. This work was supported in part by NSFC under Grant
U1908209, 62021001 and the National Key Research and Development Program
of China 2018AAA0101400.
\end{comment}

%explore one simple but effective pretraining strategy by  
%In this paper, we propose HST, a hierarchical network using swin transformer to solve the compressed image super-resolution problem. The HST consists of four parts: HFM first extracts three-level feature maps, then FEM uses residual swin transformer blocks to enhance features, and FM fuses features from different branches, HRM finally generates a clean high-resolution image. Extensive experiments demonstrate that, with pretraining and hierarchical network structure, our HST achieves best performance on compressed image super-resolution tasks. In addition, our model takes fifth place of AIM2022 challenge, with the PSNR of 23.51dB.

\clearpage
\bibliographystyle{splncs04}
\bibliography{egbib}
\end{document}